\documentclass{article}
\usepackage{spconf,amsmath,graphicx}
\usepackage{nccmath}
\usepackage[percent]{overpic}
\usepackage{color}
\usepackage{xcolor}
\usepackage{url}
\usepackage{fontawesome}

\newcommand{\ignore}[1]{}
\title{accurate 3D cell segmentation using deep feature and CRF refinement}
\name{Jiaxiang Jiang$^{\star}$  Po-Yu Kao$^{\star}$ Samuel A. Belteton$^{\dagger}$    Daniel B. Szymanski$^{\dagger}$  B.S. Manjunath$^{\star}$ \thanks{This work was supported by NSF MCB Grant No. 1121893 to D.B.S. and NSF MCB Grant No. 1715544 to B.S.M.}}

\address{$^{\star}$ Department of Electrical and Computer Engineering, University of California, Santa Barbara \\$^{\dagger}$ Department of Botany and Plant Pathology, Purdue University}
\begin{document}
%
\maketitle
\begin{abstract}
We consider the problem of accurately identifying cell boundaries and labeling individual cells in confocal microscopy images, specifically, 3D image stacks of cells with tagged cell membranes. Precise identification of cell boundaries, their shapes, and quantifying inter-cellular space leads to a better understanding of cell morphogenesis. Towards this, we outline a cell segmentation method that uses a deep neural network architecture to extract a confidence map of cell boundaries, followed by a 3D watershed algorithm and a final refinement using a conditional random field. In addition to improving the accuracy of segmentation compared to other state-of-the-art methods, the proposed approach also generalizes well to different datasets without the need to retrain the network for each dataset. Detailed experimental results are provided, and the source code is available on GitHub \faGithub\footnote{\url{https://github.com/UCSB-VRL/Purdue3DCell}}. 
\end{abstract}
\begin{keywords}
Cell segmentation, convolutional neural networks, 3D U-Net, 3D watershed, conditional random field
\end{keywords}
\section{Introduction}
\label{sec:intro}

Accurate cell boundary detection from 3D confocal imagery is critical in many applications including modeling of cell morphogenesis \cite{purdue} and plant growth \cite{merge}. 
A typical cell segmentation workflow consists of the following steps: (i) image pre-processing to enhance the signal-to-noise ratio, (ii) a segmentation method that then partitions the cells, and (iii) a final refinement step. 
Typical pre-processing methods including morphological operations, edge enhancement methods, and denoising are widely used in the recent cell segmentation tasks \cite{RACE,Morph2,GVF,filtering}. 
This pre-processing stage is often data-dependent and requires parameter tuning to avoid under- or over-segmentation. 
For membrane-tagged images, segmentation is usually carried out with either 3D watershed based methods \cite{merge,Morph2}  or 3D level sets \cite{levelset}. 
Some methods work on individual 2D images and then fuse the results to get a 3D segmentation \cite{RACE,cellect}. 
Finally, post-processing using cell volume or shape heuristic is usually applied. 
For this traditional workflow of cell segmentation, pre-processing is subjective, and its parameters are highly dependent on the data. 

During the past few years, there has been much interest in the use of deep learning for the cell segmentation problems, and the U-Net in particular has been widely explored \cite{unet2,unet3}. U-Net is a fully convolutional network which consists of encoder and decoder parts.  The encoder part uses context information to encode raw images into feature maps, and the decoder part uses the produced feature maps to localize objects and generate the segmentation. U-Net based method gives fairly good cell boundary segmentation accuracy on the similar dataset but its performance drops in generalizing to other datasets. Besides, U-Net is not guaranteed to generate closed cell surfaces.

\begin{figure}[ht] 
	\centering
	\begin{overpic}[width=0.216\textwidth]{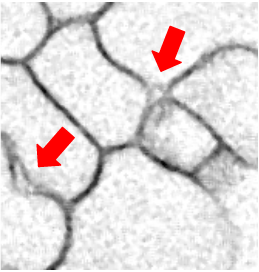}\put(2,2){A}\end{overpic}
	\begin{overpic}[width=0.216\textwidth]{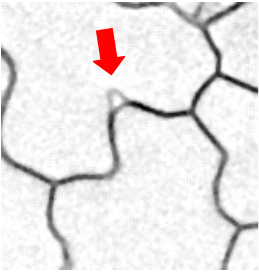}\put(2,2){B}\end{overpic}
    \caption{(A) Inter-cellular spaces and (B) \textit{Protrusion} are indicated by red arrows. \textbf{Best viewed in color.}}
    \label{fig:example}
\end{figure}

In this paper, we aim to solve the problem of accurately identifying cell boundaries and labeling individual cells in confocal microscopy image stacks. The goal is to generate closed cell surfaces in the 3D image stack while being able to accurately delineate features of interest such as the inter-cellular spaces and \textit{protrusions}, see fig.~\ref{fig:example}. The first step is to generate a membrane probability map where voxels that are likely to belong to a cell membrane have higher values than those that are not on the cell membrane. This is achieved by a 3D U-Net architecture which is trained on a large, public data. This is followed by a watershed method that labels individual cells. One challenge for watershed algorithm is automated seeding of the cells, and this is achieved using a simple distance transform of the membrane probability map. The watershed based approach is sensitive to the image signal and likely to smooth out the boundaries and also miss smaller features that are significant in modeling cell morphogenesis. For this purpose, we introduce a final processing step based on conditional random fields (CRFs). The CRF refines the watershed boundaries taking into account the probability map and local voxel boundary information. As demonstrated through the extensive experiments, this would help the overall approach to be sensitive to local boundary perturbation, including detecting salient boundary features. The proposed method is validated in two datasets, and the experimental results shows that our segmentation method outperforms the current cell segmentation methods \cite{merge,fernandez2010imaging,mosaliganti2012acme} with less computation time.  In addition, we also demonstrate that the proposed method generalizes well to different datasets without addition retraining or fine tuning the parameters of the neural network. 

To summarize, the main contributions of this paper include (i) a method that is sensitive to local image features, for  accurate detection and localization of boundaries in 3D confocal stacks; (ii) and show that the method is capable of generalizing to other datasets without further training.

\section{Material and Methods}
\label{sec:model}
\subsection{Dataset}
Two 3D confocal image stack datasets of fluorescent-tagged plasma-membrane cells are used in this paper. In both datasets, only the plasma-membrane signal is available and is represented by voxels with high intensity values.  
The first dataset contains 124 image stacks of cells in the shoot apical meristem of 6 \textit{Arabidopsis thaliana} \cite{data}. 
In each image stack, there are 150 to 300 slices containing of 3 layers of cells: outer layer (L1), middle layer (L2), and deep layer (L3), and the size of each slice is $512\times 512$.
The available resolution of each image in x and y direction are 0.22$\mu m$ and in z is about 0.27$\mu m$. 
We should note that as we go from L1 towards L3, the image quality degrades progressively due to scattering of light. 
This condition makes the segmentation problem challenging in L3.
In this dataset, the ground truth voxel-wise cell labels are provided, and each cell has a unique label.
The second dataset \cite{purdue} consists of a long-term time-lapse from \textit{A. thaliana's} leaf epidermal tissue that spans over a 12 hour period with a xy-resolution of 0.212$\mu m$ and 0.5$\mu m$ thick optical sections. There are 20 image stacks in this dataset. In each image stack, there are 15 to 30 slices containing one layer of cell, and the size of each slice is $512\times 512$.
The ground truth cell labels are not provided in this dataset.

\subsection{Method}
\begin{figure}[htb]
  \centering
  \centerline{\includegraphics[width=\columnwidth]{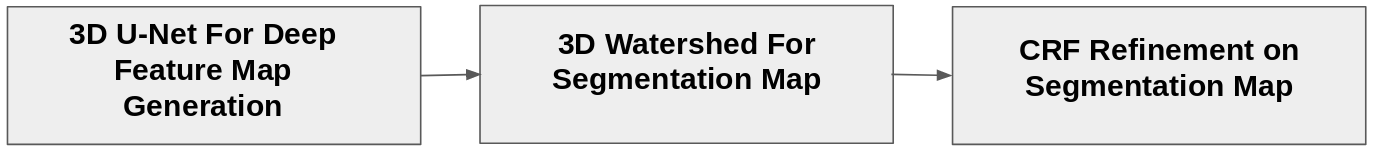}}
  \caption{The general workflow of our proposed method.}
  \label{fig:workflow}
\end{figure}

Our method for the 3D cell segmentation contains three steps as shown in fig.~\ref{fig:workflow}.
First, a 3D U-Net based neural network is used to generate a probability map of each voxel being the membrane. 
Second, a 3D watershed algorithm whose seeds are generated automatically is applied to this probability map, and outputs the initial cell segmentation result. 
Finally, a CRF is used to refine the boundaries of this initial cell segmentation. 

\subsubsection{Pre-processing}
First, we apply a background subtraction on the raw cell image using the rolling ball algorithm \cite{rolling}. 
This algorithm estimates the intensity value of background by averaging over a large neighborhood (ball) of each voxel. 
Then, we subtract this background from the raw cell image. 
Next, spatial resolution adjustment and histogram matching are applied to this background-subtracted cell image that accommodate different spatial resolution and different voxel intensity distribution of different datasets.

\subsubsection{Membrane Probability Map using U-Net}
For probability map generation, a  3D U-Net based network is used. For training the network, we used the \textit{A. thaliana} dataset \cite{data}.  Half the dataset is used for training and the other half for testing, and we only use the L1 layer cells from training set to train the network. 
The architecture of our model is based on the modified 3D U-Net \cite{isensee2017brain} with group normalization \cite{GroupNorm}. 
In our model, the input are the sub-slices of 3D stack images (of dimensions $512\times512\times16$). Due to computational limitations, we are able to input only 2 batches at a time. 
With this small batch size, standard batch normalization is unstable and tends to have a higher error so we use group normalization instead \cite{kao2018brain}.
In our implementation, the default number of groups is set to 4.
We use the Huber loss function for training, defined as:
\begin{equation} \label{eq:loss}
L_\delta(y,\hat{y}(I))=\begin{cases}\frac{1}{2}(y-\hat{y}(I))^2, \textrm{if} \  |y-\hat{y}(I)|<\delta\\ \delta|y-\hat{y}(I)|-\frac{1}{2}\delta^2, \textrm{otherwise}\end{cases}
\end{equation}
where $y\in \{ 0,1 \}$ is ground truth label representing whether the voxel is membrane, $I$ is the input image, and $\hat{y}\in [0,1]$ is the regression function which is learned by the 3D U-Net. 
The reason we use Huber loss is that the network is used to learn the regression function but not the classification function.
This loss function is quadratic when the predicted output is close to the ground truth and linear when the prediction is far from the ground truth. $\delta$ determines the threshold value between quadratic and linear loss.
This loss function is differentiable compared to a mean absolute loss function and less sensitive to outliers than a mean squared loss function. Fig.~\ref{fig:deepfeaturemap_watershed}(B) shows an example of the membrane probability map generated by the 3D U-Net.

\subsubsection{3D Watershed Segmentation}
A standard 3D watershed segmentation algorithm is now applied to the 3D probability map. 
The challenging part of the watershed segmentation is to find seeds of each cell.
In most cell images with a stained or fluorescent-tagged nucleus, the nuclei can be used as seeds. 
However, in surface labeled cell images, nuclei information is not tagged and the seed points need to be automatically generated. 

\begin{figure}[ht] 
	\centering
	\begin{overpic}[width=0.155\textwidth]{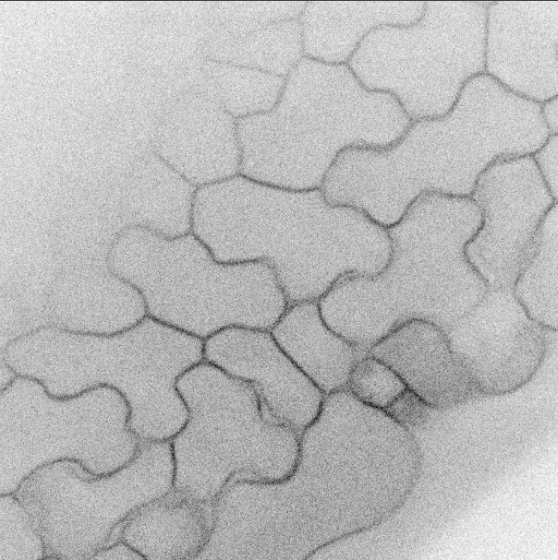}\put(90,2){A}\end{overpic}
	\begin{overpic}[width=0.155\textwidth]{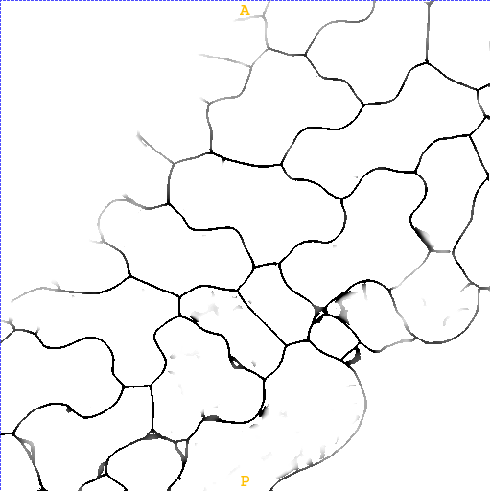}\put(90,2){B}\end{overpic}
	\begin{overpic}[width=0.155\textwidth]{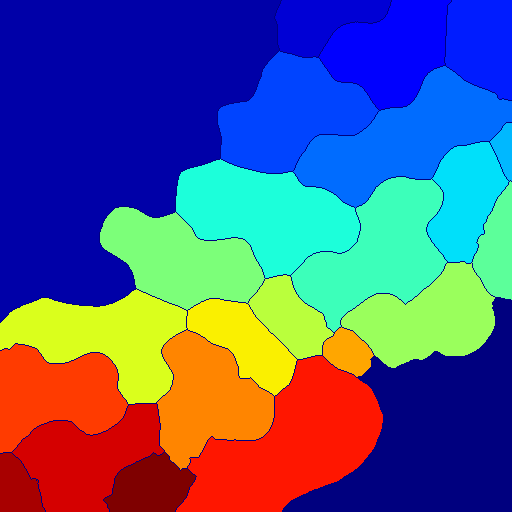}\put(90,2){\color{white}C}\end{overpic}
    \caption{(A) Inverted raw image in xy orientation, (B) inverted probability map from the 3D U-Net, (C) initial segmentation result from 3D watershed. \textbf{Best viewed in color.}}
    \label{fig:deepfeaturemap_watershed}
\end{figure}

Towards this, we use a standard Otsu's thresholding on the probability map to get a binary image, and then compute a 3D distance transform on this binary image. 
The distance transform gives the minimum distance with respect to membrane for each voxel. 
In order to generate one seed within each cell, a H-maxima transform \cite{soille2013morphological} is applied. 
The H-maximum transform suppresses all maxima in this distance transform map whose value is less than a threshold H compared to surrounding voxel values. The remaining maxima are the seed locations, to which the 3D watershed method is applied.
Fig.~\ref{fig:deepfeaturemap_watershed}(C) shows the result of 3D watershed segmentation. 

\subsubsection{CRF Refinement}

\begin{figure}[ht] 
	\centering
	\begin{overpic}[width=0.155\textwidth]{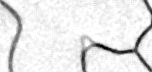}\color{black}\put(10,3){A}\end{overpic}
	\begin{overpic}[width=0.155\textwidth]{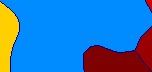}\color{white}\put(10,3){B}\end{overpic}
	\begin{overpic}[width=0.155\textwidth]{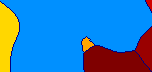}\color{white}\put(10,3){C}\end{overpic}
    \caption{(A) Inverted probability map from the 3D U-Net, (B) initial segmentation without CRF refinement, (C) final segmentation result with CRF refinement. \textbf{Best viewed in color.}}
    \label{fig:crf}
\end{figure}

In the final step we use a dense CRF \cite{crf} to refine the segmentation boundaries. 
For the given probability map \textbf{Q} = $\hat{y}(I)$ from eq.~\ref{eq:loss} and initial cell labels \textbf{X}, the conditional random field is modeled by the Gibbs distribution,
\begin{equation*}
P(\textbf{X}|\textbf{Q})=\frac{1}{Z(\textbf{Q})}\exp(-E(\textbf{X}|\textbf{Q}))
\end{equation*}
where denominator $Z(\textbf{Q})$ is the normalization factor. 
The exponent is the Gibbs energy function (for notation convenience, all conditioning is omitted from this point for the rest of the paper) and we need to minimize the energy function $E(\textbf{X})$ to get the final refined label assignments. 
In the dense CRF model, the energy function is defined as 
\begin{equation} \label{eq:energy}
E(\textbf{X})=\sum_i\psi_u(x_i)+\sum_{i<j}\psi_p(x_i,x_j)    
\end{equation} 
where $x_i$ and $x_j$ are the vertices of CRF, and $i$ and $j$ are the indices of each voxel which iterate over all voxels in the graph.
$i,j\in \{ 1,2,...,N \}$ and N is the total number of voxels in the image stack.
$x_i,x_j\in \{ 1,2,...,L \}$ and L is the total number of cells (seeds) identified by the watershed method. 
The first term of eq.~\ref{eq:energy}, the unary potential, is given by
$\psi_u(x_i)=-\log{P(x_i)},$
where $P(x_i)$ is the probability of voxel $i$ having the specific label $x_i$. It is calculated based on the probability map and the label image of the watershed. $P(x_i)=1-q_i$ if voxel $i$ is inside the cell with label $x_i$ after the watershed, and $P(x_i)=0$ otherwise. 
$q_i$ is the $i_{th}$ voxel value in the probability map. $1-q_i$ represents the probability of voxel being the interior point of the cell.
The pairwise potential in eq.~\ref{eq:energy} takes into account the label of neighborhood voxels to make sure the segmentation label is closed and the boundary is smooth. 
It is given by:
\begin{equation*}
    \psi_p(x_i,x_j)=\mu(x_i,x_j)\sum_mw^{(m)}k^{(m)}(\textbf{f}_i,\textbf{f}_j)
\end{equation*}
 where the penalty term $\mu(x_i,x_j)=1$ if $x_i\neq x_j$, and $\mu(x_i,x_j)=0$ otherwise. $w^{(m)}$ is the weight for each segmentation label $m\in \{1,2,...,L\}$, and $k^{(m)}$ is the pairwise kernel term for each pair of voxels $i$ and $j$ in the image stack regardless of their distance that capture the long-distance voxel dependence in the image stack. $\textbf{f}_i$ and $\textbf{f}_j$ are feature vectors from the probability map. $\textbf{f}_i$ incorporates location information of voxel $i$ and the corresponding value in the probability map ($\textbf{f}_i = <\textbf{p}_i,q_i>$ where $\textbf{p}_i = <x_i, y_i, z_i>$, and $x_i, y_i$ and $z_i$ are the voxel $i$ in $x$, $y$, and $z$ coordinate). Specifically, the kernel $k$ is defined as
 \begin{equation*}
     \medmath{w_1\exp(-\frac{||\textbf{p}_i-\textbf{p}_j||^2}{2\sigma^2_\alpha}-\frac{||q_i-q_j||^2}{2\sigma^2_\beta})+w_2\exp(-\frac{||\textbf{p}_i-\textbf{p}_j||^2}{2\sigma^2_\gamma}})
 \end{equation*}
where the first term depends on voxel location and the corresponding voxel value in probability map. The second term only depends on the voxel location. $\sigma_\alpha$, $\sigma_\beta$, and $\sigma_\gamma$ are the hyper parameters that depends on shape of cells in each dataset.
Fig.~\ref{fig:crf} shows an example of the results before and after CRF refinement.

\begin{figure*} [t!]
\centering
\begin{overpic}[width=0.16\textwidth]{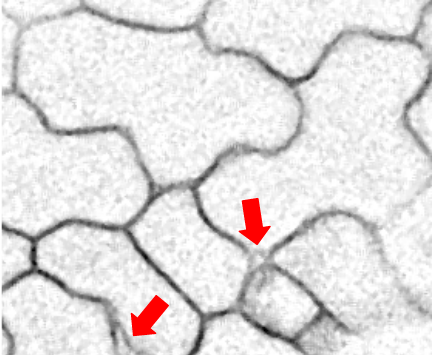}\put(5,3){A}\end{overpic}
\begin{overpic}[width=0.16\textwidth]{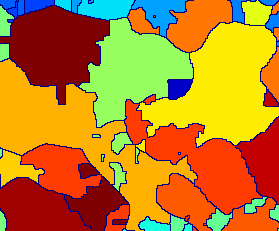}\put(5,3){B}\end{overpic}
\begin{overpic}[width=0.16\textwidth]{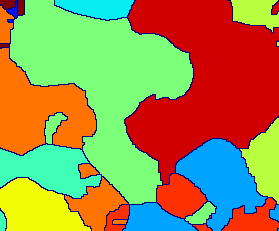}\put(5,3){C}\end{overpic}
\begin{overpic}[width=0.16\textwidth]{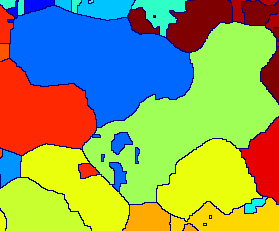}\put(5,3){D}\end{overpic}
\begin{overpic}[width=0.16\textwidth]{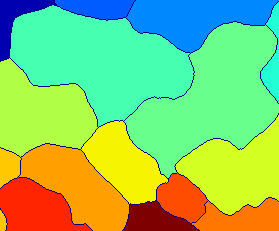}\put(5,3){E}\end{overpic}
\begin{overpic}[width=0.16\textwidth]{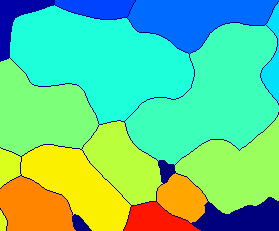}\put(5,3){F}\end{overpic}\vspace{0.5mm}
\begin{overpic}[width=0.16\textwidth]{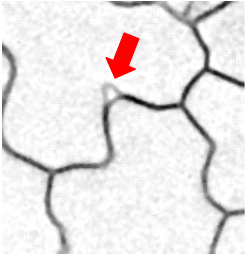}\put(5,3){A}\end{overpic}
\begin{overpic}[width=0.16\textwidth]{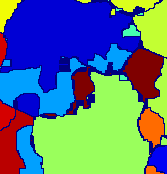}\put(5,3){B}\end{overpic}
\begin{overpic}[width=0.16\textwidth]{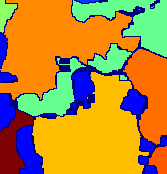}\put(5,3){C}\end{overpic}
\begin{overpic}[width=0.16\textwidth]{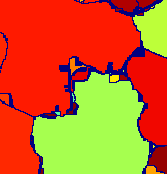}\put(5,3){D}\end{overpic}
\begin{overpic}[width=0.16\textwidth]{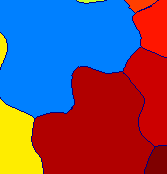}\put(5,3){E}\end{overpic}
\begin{overpic}[width=0.16\textwidth]{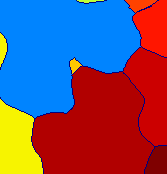}\put(5,3){F}\end{overpic}
\caption{The top row shows the segmentation results of the cell image with inter-cellular spaces indicated by red arrows and the bottom row shows the segmentation results of the cell image with a \textit{protrusion} pointed by a red arrow. (A) Inverted raw image in xy orientation, (B) MARS, (C) ACME, (D) supervoxel-based method, (E) proposed method without CRF, (F) proposed method. \textbf{Best viewed in color.}}
\label{fig:result2}
\end{figure*}

\section{Experimental Results}
\label{sec:result}
We apply our cell segmentation method on 3D confocal image stacks of \textit{A. thaliana} shoot apical meristem cells \cite{data}. 
The $L_1$ layer of the training set is used to train the 3D U-Net, and $L_1$, $L_2$ and $L_3$ layers of the testing set are used to evaluate the segmentation performance of the proposed algorithm. 
Table~\ref{tab:l1} to \ref{tab:l3} show the comparison of the final result using our proposed method and other methods including ACME \cite{mosaliganti2012acme}, MARS \cite{fernandez2010imaging} and a supervoxel-based algorithm \cite{merge} on L1 to L3 respectively. 
With CRF refinement, our model improves approximately 0.028 in the F-score measure on average. 
In addition, our proposed model has the best F-score measure on L1, L2 and L3 compared to other methods. 
It is noted that the average segmentation time of our proposed model is significantly shorter compared to the supervoxel-based method \cite{merge}.
Our proposed method takes approximately 0.6 seconds to segment one slice of image stacks on average, whereas supervoxel-based method takes approximately 6 seconds on a NVIDIA GTX Titan X and an Intel Xeon CPU E5-2696 v4 @ 2.20GHz. 
\begin{table}[ht]
\centering
\caption{3D Segmentation Performance on L1}
\label{tab:l1}
\begin{tabular}{|c|c|c|c|}
    \hline
     Algorithm & Precision&Recall&F-Score \\
    \hline
     ACME&0.805 &\textbf{0.966} &0.878\\
     MARS&0.910 &0.889 &0.899\\
     Supervoxel method&\textbf{0.962} &0.932 &0.947\\
     Result before CRF&0.944 &0.930 &0.937\\
     our method&0.953 &0.973 &\textbf{0.963}\\
     \hline
\end{tabular}
\centering
\caption{3D Segmentation Performance on L2}
\label{tab:l2}
\begin{tabular}{|c|c|c|c|}
    \hline
     Algorithm & Precision&Recall&F-Score \\
    \hline
     ACME&0.775 &\textbf{0.980} &0.866\\
     MARS &0.921 &0.879 &0.900\\
     Supervoxel method&0.910 &0.932 &0.921\\
     Result before CRF&0.924 &0.920 &0.922\\
     our method&\textbf{0.943} &0.973 &\textbf{0.953}\\
     \hline
\end{tabular}
\centering
\caption{3D Segmentation Performance on L3}
\label{tab:l3}
\begin{tabular}{|c|c|c|c|}
    \hline
     Algorithm & Precision&Recall&F-Score \\
    \hline
     ACME&0.745 &\textbf{0.976} &0.845\\
     MARS &0.909 &0.879 &0.894\\
     Supervoxel method&\textbf{0.982} &0.881 &0.929\\
     Result before CRF&0.933 & 0.888&0.910\\
     our method&0.943 &0.932 &\textbf{0.937}\\
     \hline

\end{tabular}
\end{table}

In the second experiment, we apply our proposed method to the second dataset \cite{purdue} for the purpose of identifying and analyzing cells based on the segmentation. 
The segmentation results of our proposed method and other current methods are shown in fig.~\ref{fig:result2}. 
Although the quantitative results are not available in this dataset, our proposed method visually has better segmentation performance, and our method is able to identify the inter-cellular spaces and \textit{protrusions} in the 3D cell image stack. 
For these image stacks, the average number of cells in the image is 25.20$\pm$10.20, and the average detected cells using our proposed method is 25.10$\pm$10.30 which is better than MARS (35.10$\pm$15.29), ACME (20.06$\pm$8.30) and the supervoxel method (27.78$\pm$11.30). 
For the cell counting task, CRF refinement does not change the number of the cells since it only modifies the boundary of the cell.
However, with CRF refinement, we are able to get more accurate cell boundaries. 

\section{Conclusion}
\label{sec:conclusion}
We present a probability map based 3D cell segmentation method which requires very few parameters for membrane tagged images.
This method contains a probability map generation, a 3D watershed transform, and a CRF refinement. 
Our experiments used  pavement cell data \cite{purdue} and \textit{A. thaliana} \cite{data} dataset, and quantitative results show that the proposed method achieves better segmentation performance compared to other current methods with much less computation time.


\bibliographystyle{IEEEbib}
\bibliography{strings,refs}

\end{document}